\documentclass[letterpaper]{article} 
\usepackage{aaai2026}  
\usepackage{times}  
\usepackage{helvet}  
\usepackage{courier}  
\usepackage[hyphens]{url}  
\usepackage{graphicx} 
\usepackage{natbib}  
\usepackage{caption} 
\usepackage{algorithm}
\usepackage{algorithmic}
\usepackage{booktabs}
\usepackage{multirow}
\usepackage{amssymb}
\usepackage{amsmath}
\usepackage{newfloat}
\usepackage{listings}
\DeclareCaptionStyle{ruled}{labelfont=normalfont,labelsep=colon,strut=off} 
\floatstyle{ruled}
\newfloat{listing}{tb}{lst}{}
\floatname{listing}{Listing}
\title{DehazeGS: Seeing Through Fog with 3D Gaussian Splatting}
\author{
    Jinze Yu\textsuperscript{\rm 1},
    Yiqun Wang\textsuperscript{\rm 1}\thanks{Corresponding author.},
    Aiheng Jiang\textsuperscript{\rm 1},
    Zhengda Lu\textsuperscript{\rm 2},
    Jianwei Guo\textsuperscript{\rm 3,\rm 4},\\
    Yong Li\textsuperscript{\rm 1},
    Hongxing Qin\textsuperscript{\rm 1},
    Xiaopeng Zhang\textsuperscript{\rm 4}
}
\affiliations{
\textsuperscript{\rm 1}College of Computer Science, Chongqing University\\
\textsuperscript{\rm 2}School of Artificial Intelligence, University of Chinese Academy of Sciences\\
\textsuperscript{\rm 3}School of Artificial Intelligence, Beijing Normal University\\
\textsuperscript{\rm 4}MAIS, Institute of Automation, Chinese Academy of Sciences

}

\usepackage{bibentry}

\begin{document}

\maketitle

\begin{abstract}

Current novel view synthesis methods are typically designed for high-quality and clean input images. However, in foggy scenes, scattering and attenuation can significantly degrade the quality of rendering. Although NeRF-based dehazing approaches have been developed, their reliance on deep fully connected neural networks and per-ray sampling strategies leads to high computational costs. Furthermore, NeRF's implicit representation limits its ability to recover fine-grained details from hazy scenes. To overcome these limitations, we propose DehazeGS, the first physics-driven 3D Gaussian Splatting (3DGS) framework for dehazing. We adopt an explicit Gaussian representation to model fog formation via a physically consistent forward rendering process, enabling reconstruction and rendering of fog-free scenes using only multi-view foggy images as input.  
Specifically, based on the atmospheric scattering model, we simulate the formation of fog by establishing the transmission function directly on Gaussian primitives via depth-to-transmission mapping. During training, we jointly learn the atmospheric light and scattering coefficients while optimizing the Gaussian representation of foggy scenes. At inference time, we remove the effects of scattering and attenuation in Gaussian distributions and directly render the scene to obtain dehazed views. Experiments on both real-world and synthetic foggy datasets demonstrate that DehazeGS achieves state-of-the-art performance. 

\end{abstract}


\section{Introduction}

In recent years, Neural Radiance Fields (NeRF) ~\cite{mildenhall2021nerf} have leveraged deep fully connected neural networks to implicitly represent 3D scenes and achieve impressive rendering quality through differentiable volumetric rendering techniques~\cite{levoy1990efficient,max1995optical}. Subsequent works have focused on enhancing NeRF's performance, particularly in terms of training speed ~\cite{chen2022tensorf,muller2022instant} and rendering efficiency~\cite{garbin2021fastnerf,yu2021plenoctrees}. A recent work, 3D Gaussian Splatting (3DGS) ~\cite{kerbl20233d}, abandons NeRF's implicit rendering approach and instead represents scenes explicitly by converting point clouds into 3D Gaussians (ellipsoids). Thanks to its tile-based multi-threaded parallel rendering mechanism, 3DGS achieves high-quality real-time rendering.

\begin{figure}[!t]
    \centering
    {
        \includegraphics[width=1.0\linewidth]{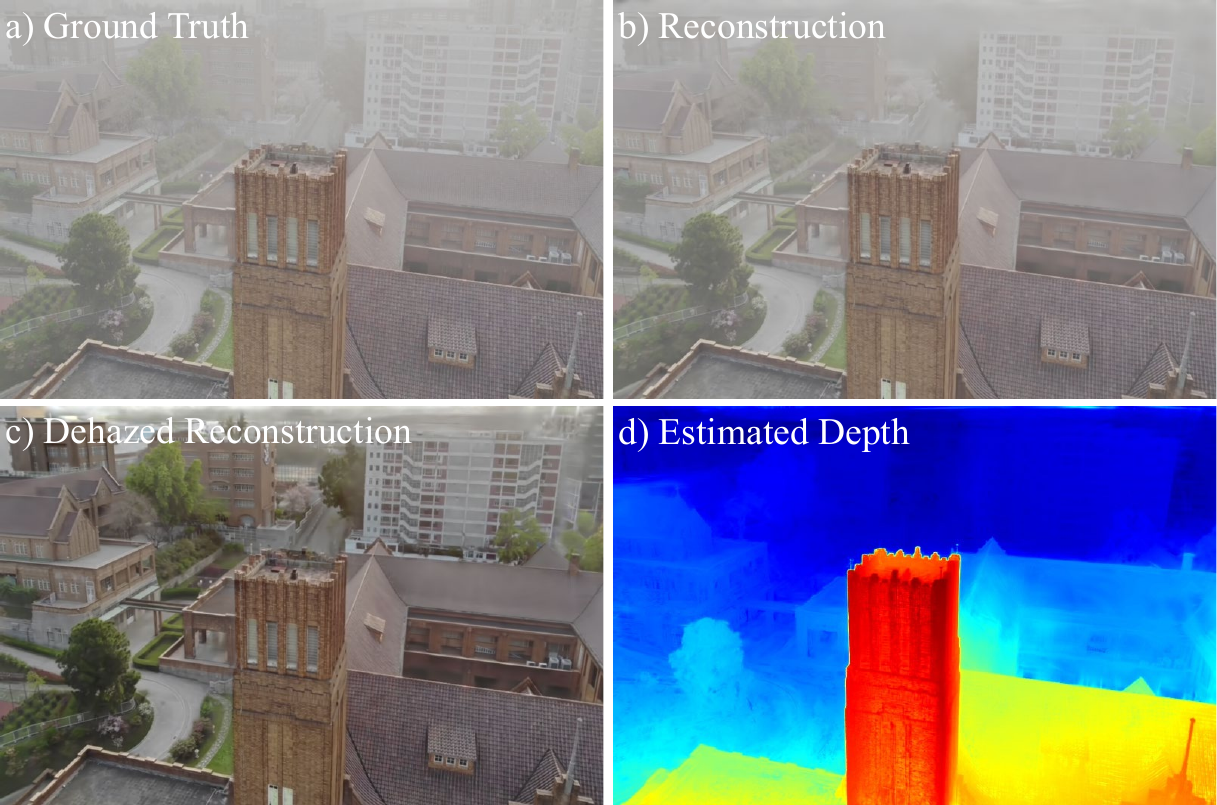}
    }
    \caption{DehazeGS can generate accurate rendering results for scenes with participating media $(b)$. By learning disentangled representations for the participating media and the underlying scene, it simultaneously recovers the clear scene $(c)$ and obtains accurate depth estimation $(d)$.}
    \label{Inpainting_flow}
\end{figure}

However, when capturing scenes containing scattering media, the light received by the detector is disrupted by the medium. The detected light mainly consists of two components: first, light reflected from the surfaces of objects that has been attenuated by particles of medium, and second, ambient light scattered by the particles of medium. This interference often leads to scenes suffering from low contrast and limited visibility. For both 3DGS and NeRF, most methods are designed for clear media, which is disadvantageous for applications such as autonomous driving and robotic operations under foggy scene conditions. Although some dehazing reconstruction methods based on NeRF ~\cite{ramazzina2023scatternerf,li2023dehazing} have been proposed, these approaches either suffer from extremely slow training and rendering speeds or are limited to specific indoor scenes. 
In contrast, traditional dehazing algorithms in low-level computer vision~\cite{zhang2018densely,yang2022self,zhang2024depth} typically require large-scale paired or unpaired datasets for training. Moreover, these methods remain confined to a 2D image plane~\cite{dong2020multi,zheng2021ultra,qin2020ffa}, lacking consideration of 3D spatial information. 
Therefore, they fail to meet the requirements of multi-view consistency across viewpoints necessary for 3D reconstruction. Feeding such inconsistent images into reconstruction models significantly degrades reconstruction quality (Detailed proofs can be found in the supplementary materials or Fig.~\ref{fig:real}). Thus, in the field of novel view synthesis, there is an urgent need for an end-to-end dehazing reconstruction model capable of effectively achieving physically accurate modeling and separating scattering effects in participating media.

To address the above challenges, we propose DehazeGS, the first differentiable physics-based 3DGS dehazing model. We have pioneered a novel representation of 3DGS for foggy scenes. 
Our method is primarily divided into three components. 1) 
Inspired by the transmission formula $t(x) = e^{- \beta d(x)}$ of the Atmospheric Scattering Model (ASM)~\cite{mccartney1976optics,narasimhan2002vision}, we propose to establish a mapping relationship between Gaussian depth $d_{\mathcal{G}}(\boldsymbol{x})$ and Gaussian transmission $t_{\mathcal{G}}(\boldsymbol{x})$.
By querying the depth of Gaussians as input to a convolutional neural network, we obtain the Gaussian transmission. The scattering coefficients are stored as the weights of the convolutional neural network. 2) After obtaining the Gaussian transmission, we model the foggy Gaussians based on the Atmospheric Scattering Model (ASM), in conjunction with the estimated atmospheric light coefficients. Once the foggy Gaussians are rasterized, the estimated fog map can be rendered. Using the real foggy map as supervision, we jointly optimize the parameters of the Gaussians while training the neural network and fully render the foggy scenes. 3) 
The accuracy of depth is crucial for estimating the transmission. Therefore, while optimizing the Gaussian representation, we adopt pseudo-depth maps as priors to impose depth regularization. To emphasize the restoration of distant details, we incorporate a depth-weighted reconstruction loss. In addition, note that for the Gaussian transmission obtained in the first stage, we directly perform alpha blending to render the transmission map. We propose to use the transmission map generated by the dark channel prior~\cite{he2010single} and the bright channel prior \cite{zhang2021single} as supervision to guide the optimization of the rendered transmission map. 

Our main contributions are as follows:
\begin{itemize}
\item We propose the first framework for learning clear 3D Gaussian splatting solely from multi-view foggy images. This framework is capable of learning disentangled representations of participating media and clear scenes, achieving fast, high-quality dehazing reconstruction.

\item We establish a mapping between Gaussian depth and its transmission, and based on the Atmospheric Scattering Model (ASM), we propose a representation method for Gaussian primitives in foggy scenes.

\item We propose a novel Gaussian-based prior loss committee, leveraging these physical priors to guide the fine-tuning of our model. The combination of these prior methods further enhances the dehazing rendering performance of our model.

\item Experiments on synthetic and real foggy datasets demonstrate that our method surpasses existing approaches in both rendering quality and speed, achieving state-of-the-art results.
\end{itemize}

\section{Related Works}
\subsection{Dehazing Based on Image Processing}
In the field of image processing, one of the most classic dehazing algorithms is the Dark Channel Prior (DCP) algorithm \cite{he2010single}, which is based on statistical observations from a large number of outdoor haze-free scenes. 
Subsequently, many other related algorithms have also been developed, such as the Bright Channel Prior \cite{zhang2021single} and the Color Attenuation Prior \cite{zhu2015fast}. 
With deep learning advancement, neural networks are increasingly applied to image processing, categorized into physics-based approaches and end-to-end methods. In physics-based algorithms, they primarily estimate hazy images' atmospheric light and transmission maps. For example, DCPDN \cite{zhang2018densely} proposed a densely connected pyramid network capable of jointly estimating atmospheric light and transmission maps. PSD \cite{chen2021psd} introduced a dehazing network guided by physical priors. For end-to-end dehazing methods, for example, DehazeNet \cite{cai2016dehazenet} utilizes a CNN to learn the mapping relationship between the original hazy images and the corresponding medium transmission maps. 
AOD-Net \cite{li2017aod} is a lightweight end-to-end dehazing network that can directly generate haze-free images without estimating transmission maps or atmospheric light independently. MSBDN \cite{dong2020multi} proposed a multi-scale enhanced dehazing network with dense feature fusion. DehazeFormer\cite{song2023vision} is designed as an end-to-end single-image dehazing method based on the Swin Transformer architecture.  
\subsection{Neural Radiance Fields for Dehazing}
NeRF \cite{mildenhall2021nerf} employs a deep fully connected neural network for scene representation, which takes 5D sampled point coordinates as input and outputs volumetric density and view-dependent radiance (i.e., color). Volumetric rendering is then used to integrate the color and density along each viewing ray to generate pixel colors.
ScatterNeRF \cite{ramazzina2023scatternerf} builds upon NeRF by introducing an additional MLP to learn the opacity of the scattering medium and its associated color. A scattering term is added to the volumetric rendering equation, blending clear and blurred sampled points to render foggy images. Its primary limitation lies in the need for high-density sampling along each ray in space, which requires processing through a deep fully connected network to compute the volumetric density and color of sampled points. DehazeNeRF \cite{chen2023dehazenerf} extends the volumetric rendering equation via simulating atmospheric scattering physics, integrating regularization strategies for dehazing and 3D shape reconstruction. However, it is mainly designed for indoor scenes. Dehazing-NeRF’s \cite{li2023dehazing} scattering coefficient/atmospheric light estimation network needs pre-training, lacking joint optimization of scattering medium and clear content. Additionally, it is confined to synthetic datasets with single objects (e.g., Lego, chair). 


\begin{figure*}[t]
    \centering
    \includegraphics[width=0.98\linewidth, height=0.46\textheight]{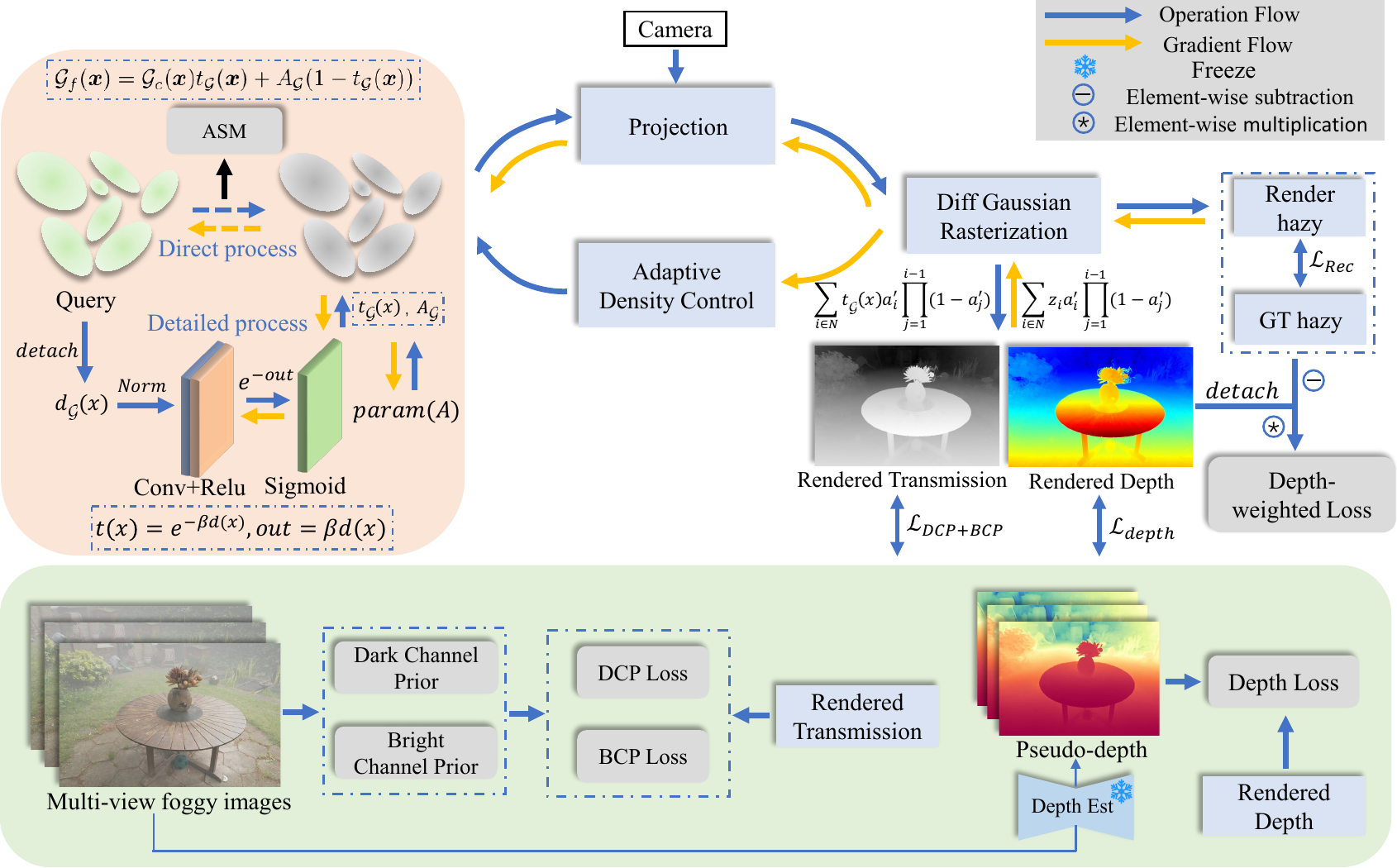}
    \caption{DehazeGS overview. We first obtain the Gaussian distributions in foggy scenes and perform alpha blending on the transmission of each Gaussian distribution to render the transmission map, which is guided and optimized using DCP and BCP priors. We utilize pseudo-depth maps as prior information for depth estimation when optimizing each input image.}
    \label{fig:pipeline}
\end{figure*}

\section{Method}\subsection{Preliminary}
\noindent \textbf{Atmospheric Scattering Model.}
The Atmospheric Scattering Model \cite{mccartney1976optics,narasimhan2002vision} (ASM) is a physical model that describes the changes in light as it propagates through the atmosphere due to scattering and absorption. Based on the principles of light propagation, it reveals how particles in the atmosphere affect the quality of captured images. Its mathematical formulation is expressed as:
\begin{equation}
    I{(x)} = J(x)t(x) + A(1 - t(x)).
\end{equation}
Here, $I$ represents the observed hazy image, $J$ denotes the underlying clear image, and $A$ is the global atmospheric light, 
which represents the background light intensity received by the imaging devices. It is generated by the scattering of ambient light within participating media and primarily affects the overall brightness of the image. $t(x)$ represents the transmission map, where $x$ denotes the position of a pixel. The transmission map primarily serves to characterize the attenuation of light as it travels through scattering media (fog) in the atmosphere, directly influencing the accuracy of the dehazing results. $t(x)$ can be further expressed as $t(x) = e^{- \beta d(x)}$, where $\beta$ represents the scattering coefficient and $d(x)$ denotes the scene depth at pixel $x$.

\noindent \textbf{3D Gaussian Splatting.}
The 3D Gaussian approach does not rely on neural radiance fields. Instead, it represents the scene as a series of 3D Gaussian distributions~\cite{yu2024spikegs}. Based on the initialized sparse point cloud, a set of 3D Gaussians, defined as $G$, is parameterized by its 3D coordinates $\boldsymbol{x} \in \mathbb{R}^{3}$, 3D covariance $\boldsymbol{\Sigma} \in \mathbb{R}^{3 \times 3}$, opacity $\alpha \in \mathbb{R}$ and color $c \in \mathbb{R}^{3}$. $c$ is represented by spherical harmonics for view-dependent appearance. 
The distribution of each Gaussian is defined as:
\begin{equation}
    {\mathcal{G}}( \boldsymbol{x} ) = e^{- \frac{1}{2}{({\boldsymbol{x} - \boldsymbol{\mu}})}^{\boldsymbol{T}}\boldsymbol{\Sigma}^{- 1}(\boldsymbol{x} - \boldsymbol{\mu})}.
\end{equation}
In order to ensure the positive semi-definite property of the covariance matrix during the optimization, it is further expressed as:
\begin{equation}
    \boldsymbol{\Sigma} = \boldsymbol{R}\boldsymbol{S}\boldsymbol{S}^{\boldsymbol{T}}\boldsymbol{R}^{\boldsymbol{T}},
\end{equation}
where $\boldsymbol{R}$ and $\boldsymbol{S}$ denote the rotation and scaling matrix.

3D gaussians are projected into the 2D image space from a given camera pose $\boldsymbol{P}_{\boldsymbol{c}} = \{ \boldsymbol{R}_{\boldsymbol{c}} \in \mathbb{R}^{3 \times 3},\boldsymbol{~}\boldsymbol{t}_{\boldsymbol{c}} \in \mathbb{R}^{3} \}$. 
Given the viewing transformation $\boldsymbol{W}$ and 3D covariance matrix $\boldsymbol{\Sigma}$, the projected 2D covariance matrix $\boldsymbol{\Sigma}'$ is computed using, as described in~\cite{zwicker2001ewa}:
\begin{equation}
    \boldsymbol{\Sigma}' = \boldsymbol{J}\boldsymbol{W}\boldsymbol{\Sigma}\boldsymbol{W}^{T}\boldsymbol{J}^{T},
\end{equation}
where $\boldsymbol{J}$ is the Jacobian of the affine approximation of the projective transformation.

Subsequently, the transformed Gaussians are sorted based on their depth, and the sorted Gaussians are rasterized to render pixel values using the volume rendering equation:
\begin{equation}
    C = {\sum\limits_{i \in N}c_{i}}\alpha'_{i}{\prod\limits_{j = 1}^{i - 1}( 1 - \alpha'_{j} )},
\end{equation}
where $c_{i}$ is the learned color and the $\alpha'_{i}$ is the multiplication result of the learned opacity $\alpha_{i}$ and the 2D projected Gaussion.
\subsection{DehazeGS}
\textbf{Modeling Gaussian distributions in foggy scenes}. Fog is a natural phenomenon caused by the scattering effect of aerosol particles in the atmosphere. When photographing in foggy conditions, the light received by the sensor primarily originates from two sources: the light reflected from the object's surface, which undergoes scattering and attenuation by the particles, along with the environmental light, which is also scattered by the particles. As light passes through the scattering medium, it experiences varying degrees of scattering and attenuation, with the attenuation increasing as the scene depth increases (i.e., the transmission rate decreases). Since the scene represented by 3DGS is modeled using 3D Gaussians, we propose directly defining the transmission function $t(x)$ on each Gaussian distribution. The physical meaning of this function is the proportion of light intensity that remains unattenuated after passing through the particles. Furthermore, the transmission function is associated with the depth of the Gaussian. For any Gaussian $\mathcal{G}(\boldsymbol{x})$, its depth $d_{\mathcal{G}}(\boldsymbol{x})$ (representing the depth $d$ of the Gaussian ellipsoid at coordinate $x$) is queried, and its corresponding transmission $t_{\mathcal{G}}(\boldsymbol{x})$ (representing the transmittance $t$ of the Gaussian ellipsoid at coordinate $x$) is obtained through a single-layer one-dimensional convolutional neural network. The mapping relationship can be expressed as $\left. F_{\beta}:{d_{\mathcal{G}}(\boldsymbol{x})}\rightarrow t_{\mathcal{G}}{(\boldsymbol{x})} \right.$, where $\beta$ represents the learnable weight, that is, the scattering coefficient. The detailed structure of the network is shown in the upper-left part of Fig.~\ref{fig:pipeline}. 

Since the $t_{\mathcal{G}}(\boldsymbol{x})$ ranges between 0 and 1, with smaller values indicating more severe light attenuation, we apply a ReLU activation function to the exponential term to ensure the non-negativity of $\beta d_{\mathcal{G}}(\boldsymbol{x})$ (the product of the scattering coefficient $\beta$ and Gaussian depth), thereby restricting the output range of the function to between 0 and 1. However, in practical experiments, we found that when ensuring the non-negativity of $\beta d_{\mathcal{G}}(\boldsymbol{x})$, feeding the result of the entire negative exponential function into a sigmoid activation function can improve the training stability of the model. Although adding a sigmoid is not strictly necessary.

Finally, based on equation $t(x) = e^{- \beta d(x)}$ from Section 3.1, we can obtain the transmission function formula based on Gaussians as follows:
\begin{equation}
    t_{\mathcal{G}}{(\boldsymbol{x})} = {\sigma( e}^{- max(0,~\beta d_{\mathcal{G}}(\boldsymbol{x}))}),
\end{equation}
the depth of the Gaussian ellipsoid $d_{\mathcal{G}}(\boldsymbol{x})$ is normalized to the range $[0,1]$, and its gradient is detached to prevent the gradients from flowing through it. This approach is designed to prevent the gradient from affecting the latent representation of clear Gaussians, while allowing for better optimization of the network. For atmospheric light estimation, it is generally assumed that the atmospheric light in foggy scenes is global. The learnable global atmospheric light parameter $A_{\mathcal{G}}$ applies to all Gaussians. Consequently, the Gaussian distribution in foggy scenes is represented as follows,
\begin{equation}
    \mathcal{G}_{f}{(\boldsymbol{x})}=\mathcal{G}_{c}(\boldsymbol{x})t_{\mathcal{G}}(\boldsymbol{x}) + A_{\mathcal{G}}(1 - t_{\mathcal{G}}(\boldsymbol{x})),
\end{equation}
where $\mathcal{G}_{f}{(\boldsymbol{x})}$ and $\mathcal{G}_{c}{(\boldsymbol{x})}$ represent the fogged 3D Gaussians and the latent clear 3D Gaussians, respectively.

This approach of directly establishing the transmission function and atmospheric light on the Gaussians effectively avoids the high sampling and rendering costs associated with NeRF-based methods~\cite{chen2023dehazenerf,ramazzina2023scatternerf}. Moreover, this approach effectively leverages the reconstruction capability in 3DGS, enhancing the structural consistency of the dehazed scene across multiple viewpoints. 

After rendering the fog image, the reconstruction loss between the rendered image $\Hat{I}$ and the real fog image $I$ is calculated. Consistent with the original 3D Gaussian reconstruction loss, The loss function is $\mathcal{L}_1$ combined with a $D-SSIM$ term, $\lambda$ = 0.2:
\begin{equation}
\mathcal{L}_{rec} = (1 - \lambda) \mathcal{L}_1 + \lambda \mathcal{L_{\textrm{D-SSIM}}}.
\end{equation}

\noindent \textbf{Gaussian Transmission Optimization.} 
We directly perform alpha blending on the transmission values $t_{\mathcal{G}}(\boldsymbol{x})$ of each Gaussian to obtain the rendered transmittance map $\mathcal{\Hat{T}}(\boldsymbol{P})$:
\begin{equation}
\mathcal{\Hat{T}}(\boldsymbol{P}) = {\sum\limits_{i \in N}t_{\mathcal{G}}(\boldsymbol{x})}\alpha'_{i}{\prod\limits_{j = 1}^{i - 1}( 1 - \alpha'_{j} )}, 
\end{equation}
where $\boldsymbol{P}$ represents pose.
We propose to use the Dark Channel Prior (DCP) \cite{he2010single} and further consider Bright Channel Prior (BCP)~\cite{sun2016image} algorithms to guide the optimization of our estimated transmission map. The DCP is one of the most well-known dehazing algorithms, leveraging the statistical properties of clear images to estimate the transmission map and atmospheric light. we follow the prior of traditional methods \cite{chen2021psd,golts2019unsupervised} and reformulate the prior as a 3DGS-based energy function:
\begin{equation}
    \begin{split}
        \mathcal{L}_{DCP} &= {{\mathcal{{T}}}_{D}}^{T}(\boldsymbol{P})L\mathcal{{T}}_{D}(\boldsymbol{P}) + \lambda{\tilde{\mathcal{T}}(\boldsymbol{P})}^{T}\tilde{\mathcal{T}}(\boldsymbol{P}), \\
        \tilde{\mathcal{T}}(\boldsymbol{P}) &= (\mathcal{T}_{D}(\boldsymbol{P}) - \mathcal{\Hat{T}}{(\boldsymbol{P})}). \\
    \end{split}
\end{equation}
Here, $\mathcal{T}_{D}(\boldsymbol{P})$ represents the transmission map under the corresponding pose $\boldsymbol{P}$, estimated by the DCP algorithm. $L$ and $\lambda$ are the Laplacian-like matrix and hyper-parameter, respectively. 

To mitigate the darkening effect caused by DCP, we further introduce BCP, a brightness enhancement algorithm based on the bright channel prior, which helps to enhance the overall brightness of the rendered dehazed images. The loss function is formulated as follows,
\begin{equation}
    \left.\mathcal{L}_{BCP} =~\middle| \middle| \mathcal{T}_{B}(\boldsymbol{P})\left. - \mathcal{\Hat{T}}{(\boldsymbol{P})} \middle| \right|_{1} \right.
\end{equation}
where $\mathcal{T}_{B}(\boldsymbol{P})$ is estimated from the BCP.

\noindent \textbf{Depth Supervision Loss.} In real-world scenarios, as the depth increases, the influence of scattering and attenuation on light becomes more significant, resulting in reduced visibility of distant details. The accuracy of depth estimation is a key factor in obtaining a more precise transmission map. In 3DGS, depth maps can be rendered by performing alpha blending on the depths of Gaussians.
\begin{equation}
    \Hat{D} = {\sum\limits_{i \in N}z_{i}}\alpha'_{i}{\prod\limits_{j = 1}^{i - 1}( 1 - \alpha'_{j} )}.
\end{equation}
We leverage DepthAnything V2 \cite{yang2024depth} to predict the depth maps from foggy images, generating pseudo-depth maps to guide the optimization of the 3DGS-rendered depth. Since the pseudo-depth maps are estimated from the foggy inputs and inherently contain certain errors, the alignment between $D_{pseudo}$ and $\Hat{D}$ is relative rather than absolute. To address this, we apply weight decay during training to gradually reduce the weight of this loss. Specifically, we adopt a continuous learning rate decay schedule. The initial weight is set to 1 at step 0 and gradually reduced to 0.01 at the specified maximum step, with logarithmic interpolation applied between these steps.

\begin{equation}
     \left.\mathcal{L}_{d} =~\middle| \middle| D_{pseudo}\left. - \Hat{D} \middle| \right|_{1} \right.
\end{equation}
Furthermore, to enhance the recovery of distant details, we incorporate a depth-weighted reconstruction loss $\mathcal{L}_{d_{rec}}$:
\begin{equation}
    \left. \mathcal{L}_{d_{rec}} = ~\middle| \middle| \Hat{D}_{detach}\left. \cdot(\Hat{I}-I) \middle| \right|_{1} \right.
\end{equation}
Finally, our total loss is formulated as follows:
\begin{equation}
    \mathcal{L} = \mathcal{L}_{rec}+\lambda_{D}\mathcal{L}_{DCP}+\lambda_{B}\mathcal{L}_{BCP}+\lambda_{d}\mathcal{L}_{d}+\lambda_{d'}\mathcal{L}_{d_{rec}},
\end{equation}
where $\lambda_{D}$, $\lambda_{B}$, $\lambda_{d}$ and $\lambda_{d'}$  are trade-off weights.
\section{Experiments}
\subsection{Experimental Setup}
\noindent \textbf{Synthetic Foggy Dataset.} 
We chose four representative scenes from the Mip-NeRF~\cite{barron2021mip} dataset, including two indoor scenes (bonsai and counter) and two outdoor scenes (garden and stump). For each scene, we employed the Depth Anything V2 model~\cite{yang2024depth} to estimate ground-truth depths of original clear images. Subsequently, we stochastically sampled scattering coefficients and global atmospheric light for each scene. Finally, haze synthesis was performed on the images based on principles of the atmospheric scattering model. Additionally, we performed dehazing tests on a driving scene from the Waymo~\cite{sun2020scalability} dataset.

\noindent \textbf{Real Foggy Dataset.} We used three indoor foggy scenes provided by DehzeNeRF \cite{chen2023dehazenerf}, captured with two professional fog machines and one mobile phone. The three scenes are bear, elephant, and lion. 
Each scene contains a certain number of clear views and foggy views.
We combined the clear and foggy views from the same scene and input them into COLMAP to obtain the sparse point cloud and corresponding poses.

\noindent \textbf{Evaluation methods.} During the training process, we only take the multi-view foggy views of the scene as input to render clear views, and compute evaluation metrics (PSNR / SSIM / LPIPS) using the rendered clear images and their corresponding real clear images.


\noindent \textbf{Baselines.} In addition to comparing with the original 3DGS~\cite{kerbl20233d}, we also compared our method with ScatterNeRF~\cite{ramazzina2023scatternerf} and DehazeNeRF~\cite{chen2023dehazenerf}. We further evaluated the reconstruction performance of combining low-level computer vision dehazing models with 3DGS. 
Specifically, we first dehazed the multi-view images using the pretrained models of Dehamer~\cite{guo2022image} or DehazeFormer~\cite{song2023vision} (as shown in the supplementary materials), which were then fed into 3DGS for reconstruction.
Furthermore, since SeaSplat~\cite{yang2024seasplat} (a real-time underwater rendering method) also has some dehazing reconstruction capabilities, we included it in the comparison as well. Since DehazeNeRF has not released its source code, its qualitative and quantitative results are both sourced from its original paper—for the Lion and Bear scenes (where DehazeNeRF lacks quantitative data), we calculated such results for corresponding viewpoints for fair comparison. Note that the real dataset views in DehazeNeRF were specifically cropped (removing scene edges). For fair comparison, we similarly cropped edges of our rendered views.


\noindent \textbf{Training Details.} 
ScatterNeRF was trained on three 4090 GPUs with a batch size of 2048, while all other models were trained on a single 4090 GPU. Remaining parameters followed the default settings. For our model, training was performed on a single GPU. The number of iterations was set to 30K for synthetic datasets and 3K for real-world datasets. $\lambda_{D}$, $\lambda_{B}$, and $\lambda_{d'}$ were all set to 0.1. The learning rate for the convolutional network was set to $1 \times 10^{-7}$, with a weight decay factor of 0.1, decaying over the corresponding number of iterations for synthetic and real scenes.

\begin{figure*}[!ht]
\centering
\begin{minipage}{\textwidth}
\begin{tabular*}{\textwidth}{c}
  \includegraphics[width=0.99\textwidth]{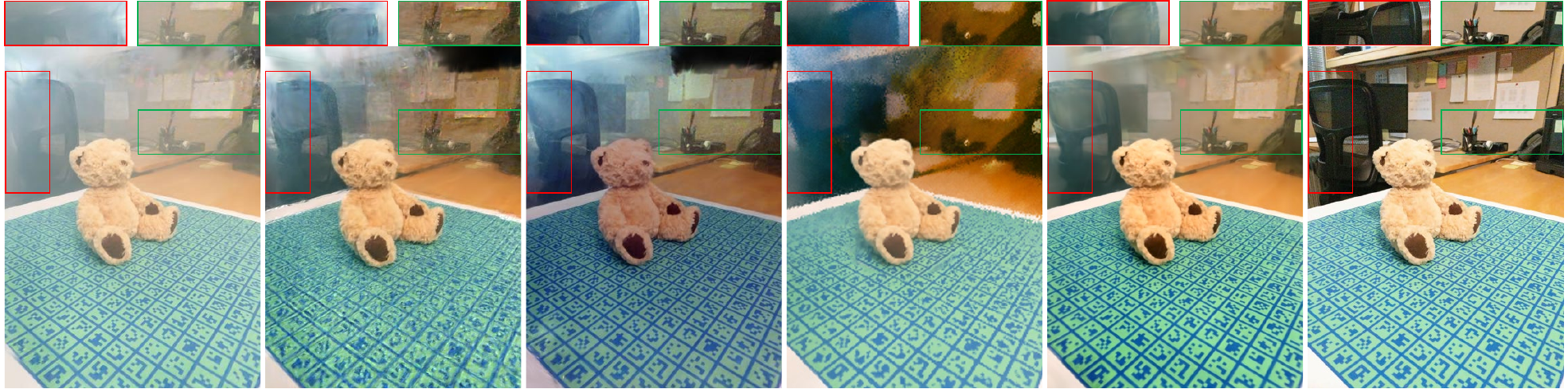}\\
  \includegraphics[width=0.99\textwidth]{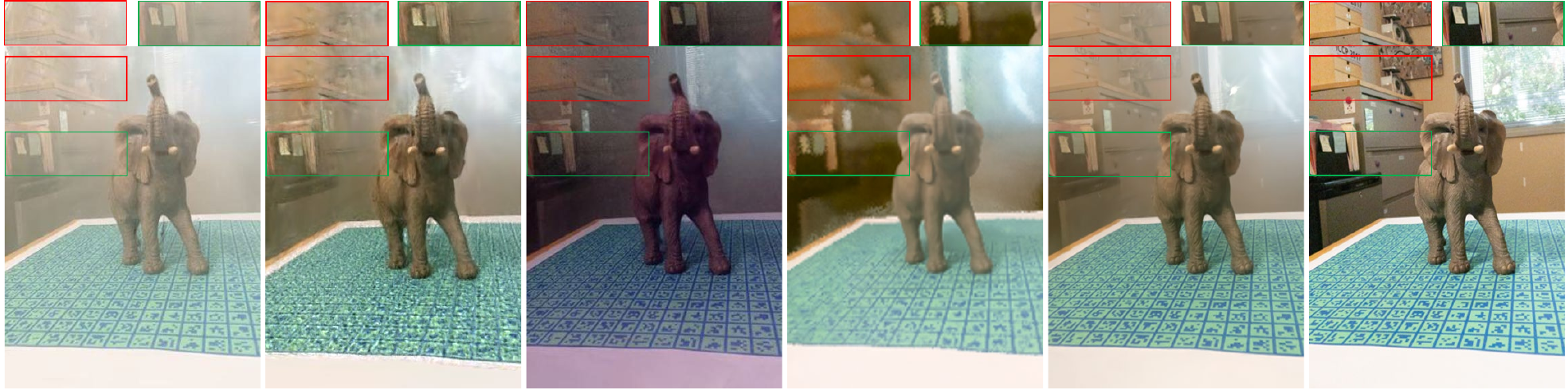}\\
  \includegraphics[width=0.985\textwidth]{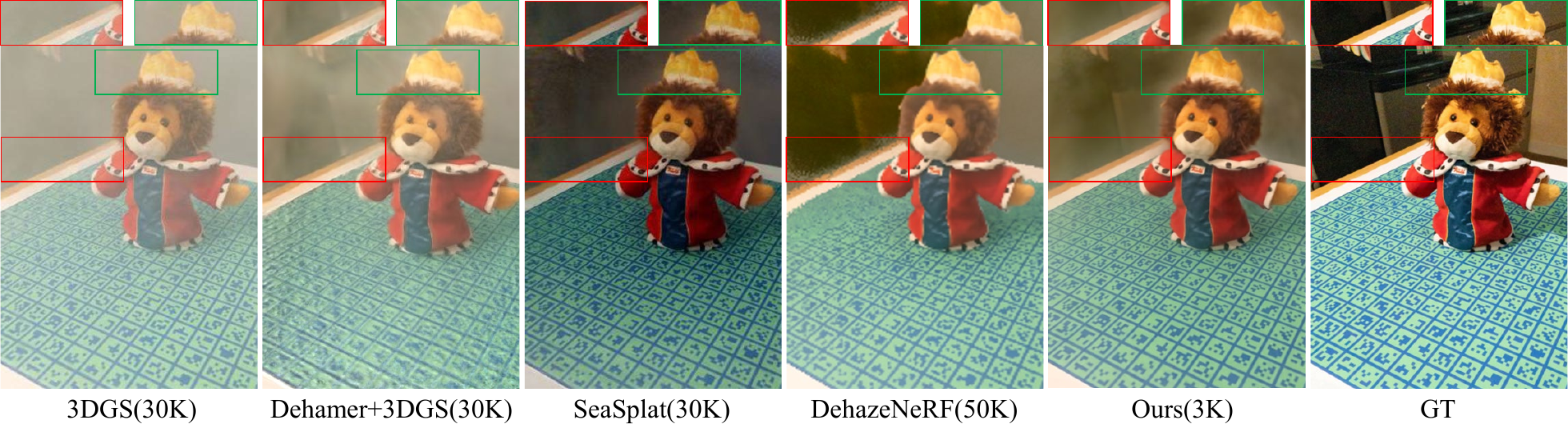}\\
  \end{tabular*}
\end{minipage}
  \caption{Qualitative comparison of novel view synthesis (input foggy images and render clear images under novel views unseen by the model) results on real datasets, the images of DehazeNeRF are taken from its original paper. We encourage readers to refer to the supplementary materials to view rendering results from more views.}
  \label{fig:real}
\end{figure*}



\begin{table*}[!ht]
\small
	\centering
	
	{
 \setlength{\tabcolsep}{3pt}
		\begin{tabular}{l|ccc|ccc|ccc|ccc|c}\toprule
			Dataset & \multicolumn{3}{c|}{ {Bear}}& \multicolumn{3}{c|}{ {Elephant}} & \multicolumn{3}{c|}{Lion}&\multicolumn{3}{c|}{ {Average}}&{\multirow{2}{*}{Train}}\\
            Method/Metric
			& PSNR & SSIM & LPIPS& PSNR & SSIM  & LPIPS& PSNR  & SSIM & LPIPS & PSNR & SSIM & LPIPS\\\midrule
            { {ScatterNeRF}} & 10.34 & 0.351 & 0.728    & 11.01 & 0.431 & 0.727 & 9.92 & 0.320 & 0.763     & 10.42 & 0.367& 0.739&$>$~16 hours\\
			{ {SeaSplat}} & 12.13 & 0.586 & \underline{0.263}    & 11.09 & 0.539 & 0.314 & 12.16 & 0.530 & 0.285 & 11.79& 0.552& 0.287&$\sim$~25 mins \\

            { {3DGS}}  & 13.09 & \underline{0.597} & 0.276     & 13.28 & 0.605 & 0.307     & 12.01 & 0.507 & 0.337 & 12.79& 0.570& 0.307& $\sim$~5 mins\\
            { {Dehamer+3DGS}}  & 13.88 & 0.393 & 0.315     & 13.36 & 0.444 & 0.368     & 13.50 & 0.431 & 0.328 & 13.58& 0.423& 0.337&$\sim$~5 mins\\
            { {DehazeNeRF}}  & \underline{14.92} &  0.439 &  0.341    &  \underline{17.87} &  \underline{0.730} &  \underline{0.150}       &  \textbf{18.01} &  \underline{0.558}
 &  \underline{0.259}  & \underline{16.93}& \underline{0.576}& \underline{0.250}& $>$~4~hours\\
            {Ours}  & \textbf{16.79} &  \textbf{0.725} &  \textbf{0.173}    &  \textbf{19.19} &  \textbf{0.791} &  \textbf{0.113}       &  \underline{17.33} &  \textbf{0.709}
 &  \textbf{0.186}  & \textbf{17.77} & \textbf{0.742} & \textbf{0.157}&$\sim$~1.2 mins\\\bottomrule
		\end{tabular}
	}
 \caption{Quantitative comparison for novel view synthesis on the real foggy dataset. 
 }
	\label{tab:real}
\end{table*}

\subsection{Quantitative and Qualitative Results}
Table~\ref{tab:real} and Table~\ref{tab:synthetic} present the quantitative comparison results on real hazy datasets and synthetic hazy datasets. Since DehazeNeRF does not provide quantitative metrics for the bear and lion scenes, we calculated them based on the rendered images presented in its paper. On real foggy datasets, our method outperforms existing methods in rendering details, with average improvements of 37.2\% (LPIPS), 28.8\% (SSIM), and 5.0\% (PSNR) over the second-best method.  For the synthetic foggy dataset, it achieves 24.3\% (LPIPS), 1.9\% (SSIM), and 4.7\% (PSNR) gains over the same baseline. 

\begin{figure*}[!t]
\centering
\begin{minipage}{\textwidth}
\begin{tabular*}{\textwidth}{c}

  \includegraphics[width=0.98\textwidth]{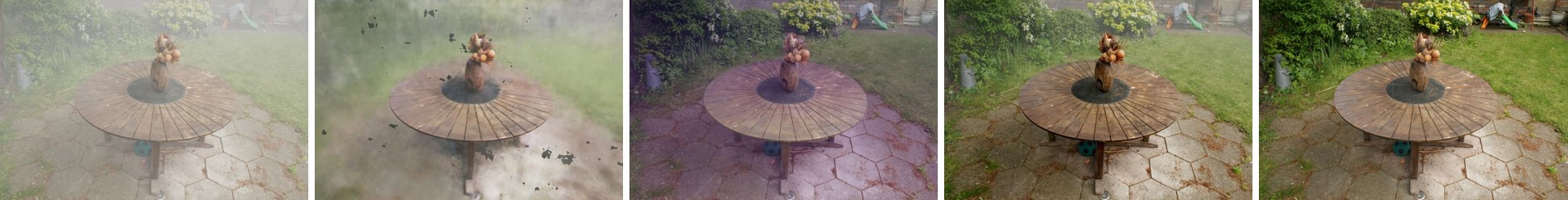}\\
  \includegraphics[width=0.98\textwidth]{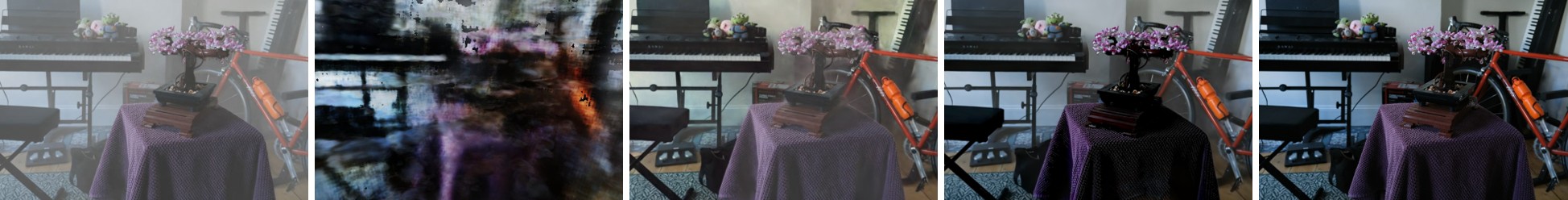}\\
  \includegraphics[width=0.98\textwidth]{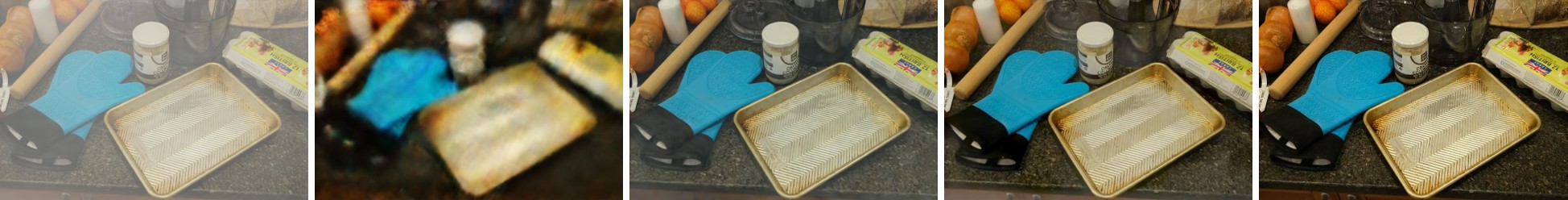}\\
  \includegraphics[width=0.98\textwidth]{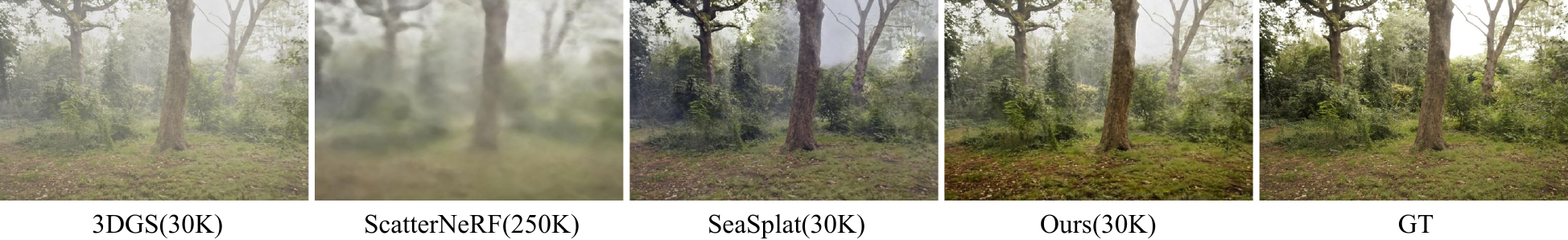}\\
  
  \end{tabular*}
\end{minipage}
  \caption{Qualitative comparison of dehazing results on the synthetic foggy dataset. Our method exhibits finer texture details and is closer to the ground truth (GT) compared to existing methods.}
  \label{fig:syn}

\end{figure*}

The qualitative comparison results are shown in Fig.~\ref{fig:real} and Fig.~\ref{fig:syn}. 
SeaSplat's dehazing results show noticeable color anomalies and missing edge regions. Compared to DehazeNeRF, our reconstruction results outperform in both dehazing quality and the details of near and far scenes. Meanwhile, ScatterNeRF fails to reconstruct clear scenes on both datasets. Additionally, our model is capable of achieving dehazing reconstruction in both indoor and outdoor complex scenarios. For additional rendering results, please refer to the supplementary materials.

\begin{table*}[!ht]
\small



	\centering
	
	{
 \setlength{\tabcolsep}{2pt}
		\begin{tabular}{l|ccc|ccc|ccc|ccc|ccc|c}\toprule
			Dataset & \multicolumn{3}{c|}{ {Garden}}& \multicolumn{3}{c|}{ {Bonsai}} & \multicolumn{3}{c|}{Counter}& \multicolumn{3}{c|}{Stump}&\multicolumn{3}{c|}{ {Average}}&{\multirow{2}{*}{Train}}\\

            Method/Metric
			& PSNR  & SSIM  & LPIPS  & PSNR  & SSIM  & LPIPS  & PSNR  & SSIM  & LPIPS & PSNR  & SSIM  & LPIPS & PSNR & SSIM & LPIPS\\\midrule
            
			{ {3DGS}} & 12.49 & 0.620 & \underline{0.255}    & 12.43 & 0.609 & 0.161 & 13.06 & 0.621 & 0.193     & 11.10 & 0.516 & 0.298     &12.27&0.592&0.227&$\sim$~10 mins\\
   { {ScatterNeRF}} & 14.15 & 0.346 & 0.609    & 18.47 & 0.492 & 0.586 & 18.32 & 0.506 & 0.633     & 11.82 & 0.224 & 0.771 &15.69
&0.392&0.649&$>$~20 hours\\
			{ {SeaSplat}}  & \underline{18.42} & \underline{0.687} & 0.275     & \underline{20.40} & \underline{0.705} & \underline{0.121}      & \underline{19.16} & \textbf{0.736} & \underline{0.183}     & \underline{17.78} & \underline{0.631} & \underline{0.293}      &\underline{18.94}&\underline{0.689}&\underline{0.218}&$\sim$~30 mins\\
            { {Ours}}  & \textbf{18.72} &  \textbf{0.742} &  \textbf{0.183}    &  \textbf{22.19} &  \textbf{0.726} &  \textbf{0.104}       &  \textbf{19.31} &  \underline{0.704}
 &  \textbf{0.153}  &  \textbf{19.10} &  \textbf{0.637} &  \textbf{0.219}  &\textbf{19.83}
&\textbf{0.702}&\textbf{0.165}&$\sim$~25 mins\\\bottomrule
		\end{tabular}
	}
 \caption{Quantitative comparison of rendering results on the synthetic fog dataset. 
 }
	\label{tab:synthetic}
\end{table*}


\subsection{Ablation Study}
In this section, we perform ablation studies on the various components of our framework to evaluate the role and contribution of each part. Specifically, we investigate the effects of depth regularization ($\mathcal{L}_{d}$), depth-weighted ($\mathcal{L}_{d_{rec}}$), and relevant physical priors ($\mathcal{L}_{DCP}$, $\mathcal{L}_{BCP}$) on rendering results. The quantitative results are presented in Table \ref{tab:Ablation}. Note that Vanilla 3DGS~\cite{kerbl20233d} refers to the original and unmodified 3DGS, without the ASM model.
(the results of its novel view synthesis are shown as 3DGS in Fig.~\ref{fig:real} and Fig.~\ref{fig:syn}).
Starting from $\mathcal{L}_{rec}$, which corresponds to our framework with only the reconstruction loss retained and all other loss terms removed, we observe that the performance progressively improves as additional losses are incorporated. 


\begin{figure}[ht]
    \centering
    \includegraphics[width=1.0\linewidth]{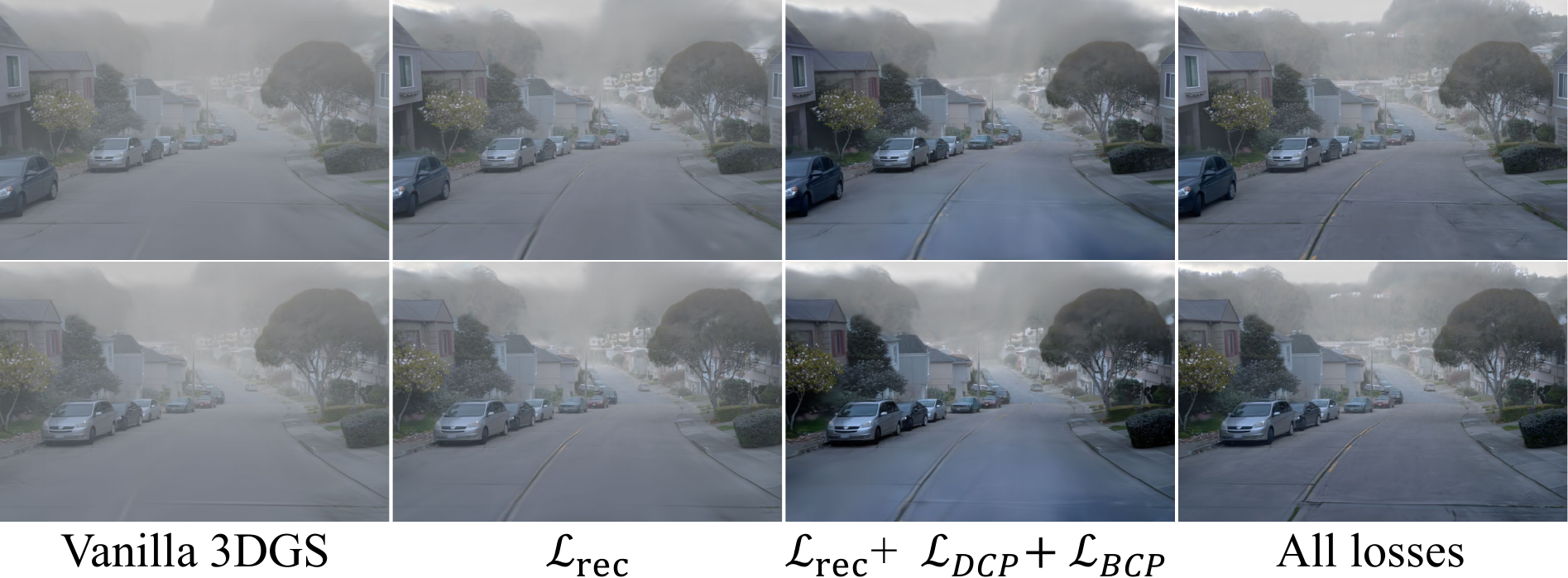}
    \caption{
    Qualitative ablation results show progressive improvements from vanilla GS to the complete model, with enhanced details and dehazing in distant and nearby scenes.}
    \label{fig:Ablation}
\end{figure}



  



\begin{table}[!ht]

\resizebox*{\linewidth}{!}{
\begin{tabular}{lccc}\toprule
{} & PSNR & SSIM  & LPIPS\\\midrule
Vanilla 3DGS & 12.27 & 0.591& 0.227 \\
$\mathcal{L}_{rec}$~(Physical Scattering Model) & 19.35 & 0.675& 0.189 \\
$\mathcal{L}_{rec} +\mathcal{L}_{DCP}$ & 19.66 & 0.687 & 0.173 \\
$ \mathcal{L}_{rec} + \mathcal{L}_{DCP} + \mathcal{L}_{BCP}$ & 19.74& \underline{0.697}& 0.175 \\
$ \mathcal{L}_{rec} + \mathcal{L}_{DCP} + \mathcal{L}_{BCP} + \mathcal{L}_{d}$  &\underline{19.77} & 0.692 & \underline{0.166} \\
All losses  & \textbf{19.83} & \textbf{0.702} & \textbf{0.164} \\\bottomrule
\end{tabular}
}
\caption{Ablation study on the synthetic foggy dataset.}
\label{tab:Ablation}
\end{table}

\section{Conclusion}
We propose the first framework for learning clear 3D Gaussian splatting solely from multi-view hazy images. Our approach leverages explicit Gaussian representation, based on the atmospheric scattering model, to interpret the formation of hazy images through a physically accurate forward-rendering process. This enables joint optimization of 3D scene representation while learning the participating medium. Overall, our method outperforms existing methods in terms of rendering quality, training speed, and rendering efficiency. However, in certain scenarios, the dehazing effect on distant scenes still needs further improvement. We hope that our approach can inspire future research in this field.

\section{Acknowledgements}
This research was supported by the National Natural Science Foundation of China (62202076), the National Key R\&D Program of China under Grant 2022ZD0160804, National Natural Science Foundation of China (U21A20515, 52238003, U22B2034, 82174224), Shenzhen S\&T programme (No. CJGJZD20240729141906008).

\end{document}